# A Novel Site-Agnostic Multimodal Deep Learning Model to Identify Pro-Eating Disorder Content on Social Media

Jonathan Feldman

# Abstract

Over the last decade, there has been a vast increase in eating disorder diagnoses and eating disorder-attributed deaths, reaching their zenith during the Covid-19 pandemic. This immense growth derived in part from the stressors of the pandemic but also from increased exposure to social media, which is rife with content that promotes eating disorders. This study aimed to create a multimodal deep learning model that can determine if a given social media post promotes eating disorders based on a combination of visual and textual data. A labeled dataset of Tweets was collected from Twitter, recently rebranded as X, upon which twelve deep learning models were trained and evaluated. Based on model performance, the most effective deep learning model was the multimodal fusion of the RoBERTa natural language processing model and the MaxViT image classification model, attaining accuracy and F1 scores of 95.9% and 0.959, respectively. The RoBERTa and MaxViT fusion model, deployed to classify an unlabeled dataset of posts from the social media sites Tumblr and Reddit, generated results akin to those of previous research studies that did not employ artificial intelligence-based techniques, indicating that deep learning models can develop insights congruent to those of researchers. Additionally, the model was used to conduct a time-series analysis of yet unseen Tweets from eight Twitter hashtags, uncovering that, since 2014, the relative abundance of content that promotes eating disorders has decreased drastically within those communities. Despite this reduction, by 2018, content that promotes eating disorders had either stopped declining or increased in ampleness anew on those hashtags.

# Introduction

## Background

In the last ten years, the number of deaths attributed to eating disorders has doubled, with new estimates claiming that eating disorders cause over three million deaths annually worldwide (Galmiche et al., 2019; Sukunesan et al., 2021; van Hoeken & Hoek, 2020). Recently, the Covid-19 pandemic has led to a drastic rise in eating disorder hospitalizations and diagnoses (Agostino et al., 2021; Matthews et al., 2022), which was caused by an increase in factors that induce eating disorders, such as isolation and stress, and the reduction of factors that would prevent eating disorders (Culbert et al., 2015). Previous studies have correlated the substantial growth in eating disorder pervasiveness and symptom severity to the advent of social media (Morris & Katzman, 2003; Sidani et al., 2016). Similarly, previous research has shown that the growth of eating disorder cases during the Covid-19 pandemic stemmed partially from increased social media usage, especially within the teenager and young adult demographic (Boepple & Thompson, 2016; Rosen et al., 2022).

Social media is home to many communities that share content capable of promoting body dissatisfaction (Feldhege et al., 2021; Rouleau & Von Ranson, 2011). Among these is a subset of communities that encourage the development of eating disorders and body dysmorphia, which researchers have dubbed Pro-Eating Disorder (Pro-ED) communities (Lewis et al., 2016; Sukunesan et al., 2021). Pro-ED communities, and the individuals that attune to their ideologies, often glorify malnutrition and starvation; guide dieting and other weight loss methodologies; and motivate weight reduction through provocative imagery called "thinspiration"—a portmanteau of thin and inspiration (Oksanen et al., 2016; Sukunesan et al., 2021). Content shared by Pro-ED communities and their members is hazardous because it can induce or exacerbate eating disorders (Mento et al., 2021). This is especially true for adolescents and young adults, who are more easily influenced by such content (Borzekowski et al., 2010; Feldhege et al., 2021). Adolescents and young adults are also the age group that spends the most time on social media (Atske, 2021), furthering their risk of exposure to Pro-ED content (Sukunesan et al., 2021). Social media platforms have begun efforts to purge Pro-ED content from their sites; however, the sheer quantity of Pro-ED content and its pervasiveness in numerous social media communities have rendered these efforts ineffectual (Chancellor, Pater, et al., 2016).

Because of the increase in eating disorder cases following the Covid-19 pandemic, many studies have been conducted into mitigation strategies for Pro-ED content (Fardouly et al., 2022; Ren et al., 2022). Among them were studies examining the ability of artificial intelligence (AI) to locate Pro-ED communities, users, and posts. Several of these studies investigated the capacity of natural language processing (NLP) models—which analyze textual input—to identify Pro-ED social media posts (Benítez-Andrades et al., 2021, 2022). Additionally, one study analyzed the ability of computer vision (CV) models—which analyze input images—to identify Pro-ED posts as well (Feldman, 2023). Despite the vast research on AI and Pro-ED content, no previous study has employed AI for multimodal analysis of Pro-ED posts. And since the utilization of multiple input modalities boosts the performance of deep learning (DL) models, and a significant amount of social media posts are a combination of textual and visual data, this is a glaring constraint for AI techniques (Abdu et al., 2021; Ilias & Askounis, 2022).

## Objectives

This study had two central objectives. The first was to create a multimodal deep learning model to identify Pro-ED social media posts from textual and visual data. The second objective was to apply the developed multimodal deep learning model to posts from several social media communities to examine the extent to which Pro-ED content has spread and how its prevalence in those communities has changed over time.

# Literature Review

Thanks to the plenitude of publicly available data on social media sites, supervised machine learning (ML) and DL techniques—which generate an output, often a prediction or classification, based on input data—have flourished. All supervised learning methods require a labeled dataset for training and testing (Melrose, 2015). The appeal of supervised ML techniques is that they can process vast inputs incredibly quickly—at a rate that no human could match— and with consistent accuracy (Goh et al., 2020; Gupta et al., 2022).

It is because of these properties—their speed and accuracy—that DL models have been applied to social media research, including studies on Pro-ED content (Benítez-Andrades et al., 2021). A previous study developed a dataset of over two thousand labeled Tweets which were then used to train several DL and ML NLP models (Benítez-Andrades et al., 2022). In an analogous manner, another study generated a dataset of over thirteen thousand Twitter posts which were then used to train several DL image classification models (Feldman, 2023). Both studies created models that attained high accuracies on their respective test datasets. Another study examined the potential risks and benefits of ML when applied to eating disorder research, determining that such models could prove highly effective (Fardouly et al., 2022). Several other studies surveyed Pro-ED social media content through a qualitative lens, analyzing the social impacts of such posts (Gordon, 2000; Khosravi, 2020).

In essence, previous studies have successfully created unimodal DL models capable of detecting Pro-ED posts on social media using NLP or CV techniques, showing that DL and ML models can recognize Pro-ED imagery and text. This study is unique because it develops and utilizes models that combine two modalities—textual and visual—to create a more robust DL model. On social media, posts are frequently a combination of textual and visual data. Thus, a model capable of analyzing both simultaneously would likely provide greater insight into the recognition of Pro-ED posts and improve the exactness of classification.

# Dataset Preparation Methods

## Data Collection

As previously stated, supervised DL and ML models require a labeled dataset from which the model can "learn." To supply such a dataset, publicly accessible Twitter posts, also called Tweets, were scraped directly from Twitter—a social platform that was recently rebranded as X. (Because this study was conducted before Twitter's rebranding, within this report, X will be addressed as Twitter for comprehensibility.) From these scraped Tweets, the content of the Tweet, and identifying information about the post, such as the Tweet identification number, any hyperlinks associated with the Tweet, and the datetime the Tweet was posted were collected. This study used Twitter to compile the dataset because previous research has documented that Twitter has a strong Pro-ED community easily identifiable by certain hashtags and keywords (Arseniev-Koehler et al., 2016; Sukunesan et al., 2021). Moreover, previous studies have shown that Pro-ED content on Twitter is largely unmoderated and is infrequently removed—despite Twitter's policy barring Pro-ED content (Sukunesan et al., 2021). Thus, Twitter is an ideal platform for a holistic analysis of unrestricted Pro-ED content, which is optimal for this study.

Based on the findings of previous studies, it was determined that the Twitter dataset would be separated into three distinct labels. The first label is the Pro-ED class, which includes posts that promote eating disorders and unhealthy eating behavior. The second label is the Pro-Recovery class, which encompasses posts promoting recovery from eating disorders, providing scientific information about eating disorders, and advocating for improving one's mental health. The third and last label is the Neutral class, which contains posts that have no explicit relation to eating disorders or the content in the Pro-ED and Pro-Recovery categories.

To generate the Pro-ED dataset, six Twitter hashtags that previous studies have shown to be hotspots for the proliferation of Pro-ED content were scraped (Arseniev-Koehler et al., 2016; Dignard & Jarry, 2021; Sukunesan et al., 2021). These hashtags were: *#proana*, *#thinspiration*, *#thinspo*, *#bonespo*, *#collarbone*, and *#thighgap*. From these hashtags, all posts from 2018 to 2022 were scraped; in total, the Pro-ED hashtags yielded 320,080 Tweets.

To compose the Pro-Recovery dataset, eight Twitter hashtags identified by previous studies as communities promoting informed eating disorder recovery, mental wellness, and positive body image were scraped using a stratified random sampling method. Random sampling was employed rather than total scraping because these hashtags are far larger than their Pro-ED counterparts (Branley & Covey, 2017; Chancellor et al., 2017; Chancellor, Mitra, et al., 2016). The hashtags used to assemble the Pro-Recovery dataset were: *#anorexiarecovery*, *#recoveryfromanorexia*, *#edrecovery*, *#eatingdisorderrecovery*, *#recovered*, *#bodypositivity*, *#selflove*, and *#edwarrior*. Altogether, the eight Pro-Recovery hashtags supplied 64,171 Tweets.

Lastly, to create the Neutral dataset, six Twitter hashtags that did not appear in any Tweet in the Pro-ED and Pro-Recovery dataset and that did not share keyword similarity with content that promotes eating disorders or content that promotes recovery from eating disorders were scraped using stratified random sampling (Chancellor et al., 2017; Counts et al., 2018; Feldman, 2023). The Neutral hashtags were: *#outfit*, *#boldmodel*, *#landscape*, *#candid*, *#photography*, and *#news*. These six Neutral hashtags ultimately provided 71,566 posts for the dataset.

## Data Cleaning

The total number of Tweets collected from the above hashtags was 455,817 and all these posts were compiled into a singular dataset. From that dataset, only posts that had both visual and textual data were isolated. Unimodal posts were removed. Afterward, any duplicate posts—duplicate posts were defined as any posts with either the same Tweet identification numbers, textual inputs, or image data— were deleted from the dataset, keeping only one copy of the post. To find duplicate images, a dHash algorithm—which compares the similarity of images based on the relationship of adjacent pixels—was used to remove any images with a similarity of over 95% (Yang et al., 2022). Finally, all hyperlinks, mentions of other Twitter users, and hashtags were removed from the text—a vital step to ensure that any ML or DL model trained on the dataset can generalize to other hashtags and social media sites. This is particularly crucial for Pro-ED content, which has been shown to escape moderation by appropriating new hashtags (Chancellor, Pater, et al., 2016; Kendal et al., 2017). After all these data sanitization steps, due to the diverse nature of social media posts, the size of the dataset shrunk to only 10,511 Tweets. Of these Tweets, 2,362 were Pro-ED, 5,020 were Neutral, and 3,129 were Pro-Recovery.

This sizeable reduction stems from the fact that most of the Tweets collected from Pro-ED hashtags were duplicates or were not multimodal in composition, thus reducing the size of the dataset. Similarly, many of the posts collected from Neutral of Pro-Recovery hashtags were not multimodal, which also led to the reduction of the dataset size.

## Data Labelling

Data labeling began first with the Pro-ED dataset. To ensure the integrity of the Pro-ED portion of the dataset, it was manually inspected per the methodologies described in previous studies (Branley &

Covey, 2017; Chancellor et al., 2017). Advertisements and non-eating disorder-related content were removed from the dataset, but overall, every effort was made to ensure that the Pro-ED content was untouched to mitigate reviewer bias (Wang et al., 2022). For the Pro-Recovery and Neutral datasets, Tweets were initially labeled based on the hashtags from which they were scraped. In other words, if the hashtag from which the post was derived was expected to be Neutral, then the Tweet was labeled Neutral.

Afterward, a vision transformer—a DL model that can analyze images—was modified to examine the similarity between images in the Pro-ED dataset and the Neutral and Pro-Recovery datasets.

In technical terminology, the methodology for image comparison was to input all images within the dataset into a pre-trained vision transformer stripped of its classification head. Without the classification head, after processing an image, the vision transformer would output a 768-dimensional embedding matrix, which could then be used to quickly compare images based on their content rather than pixel value because the vision transformer is trained to recognize image compositional elements. High dimensional matrices, however, require a great deal of storage and processing power to analyze; therefore, to minimize computation time, each matrix of embeddings was projected and then underwent a bitwise hashing, which reduced its dimensionality greatly (Johnson et al., 2017). Similar images would likely have analogous embeddings and, in turn, have similar hash values (Dubey et al., 2022). Since this dimensionality reduction method entailed a bit of randomness, multiple hash tables were maintained to avoid incidentally missing a similar image. For every image in the dataset, five similar images were found within the hash tables based on their Jaccard Similarity Index (Black et al., 2022). If the candidate images found based on the query were made up mostly of images from a class other than that of the query image, the query image was then flagged for further manual inspection. Manual inspection was necessary because comparing multidimensional embeddings is not a faultless method for image similarity, and further evaluation was requisite for proper labeling (Wan et al., 2014).

In short, all images in the dataset were compared using a DL CV model and ranked based on similarity. The five most alike images were censused. If they were of a different label than the image input into the model—the query image—then the image was flagged for further manual inspection. Ultimately, 432 images were flagged, and 373 were removed following a manual inspection of textual and visual data, leaving 10,138 Tweets in the dataset.

# Classification Methods

## Model Structure

To create a multimodal DL model that can analyze both text and images, one can either repurpose an existing multimodal model or create a new one through the fusion of two or more DL models of different modalities (Gadzicki et al., 2020). This study employed the latter technique because previous studies have shown it to create models with higher precision, recall, and accuracy (Snoek et al., 2005). Moreover, by fusing two models, one can combine state-of-the-art models from CV and NLP far more easily (Boulahia et al., 2021; Huang et al., 2020).

Within the realm of multimodal fusion, there are three main families of approach: early fusion, where information from multiple modalities is combined into one input; intermediate fusion, where the information respective to each modality is collated before classification but after input into the model; and late fusion, where the outputs of models examining different modalities are merged after classification (Boulahia et al., 2021). This study employs late fusion because it permits the use of pre-trained DL models without requiring any changes in their architecture or weights, thus preserving accuracy (Gadzicki et al., 2020). On the other hand, intermediate and early fusion require major modifications of the network architecture and or weights. And considering that fine-tuning pre-trained models provides better performance than training models ab initio, late fusion best fits the needs of this study (Sui et al., 2012).

To architect the multimodal model, this study employed a state-of-the-art tool in CV and NLP: transformers (Ruan et al., 2022; Wolf et al., 2020). A transformer is a DL model subunit that can develop contextual links between different sections of input data, providing a crucial understanding of the relationship of pieces of an input (Dosovitskiy et al., 2021). Transformers are capable of such insight because of their self-attention mechanism, which allows them to attend individual subsections of a singular input, hence the name self-attention (Lindsay, 2020).

For many years, transformers were used solely for NLP because of their faculty for understanding the relationships between words in a textual input. Recently, however, transformers have also been applied to CV, applying their self-attention mechanism to holistically analyze an image rather than relying on segmentation as Convolutional Neural Networks (CNNs) do (Cuenat & Couturier, 2022). Among the transformer models employed for NLP, the Bidirectional Encoder Representations from Transformers (BERT) has been the de facto model, and amidst the CV models, the newly developed Vision Transformer (ViT) model has become increasingly popular and is representative of the most recent advancements in image analysis (Touvron et al., 2022).

## General

Two BERT models were selected to comprise the textual processing portion of the final multimodal model: RoBERTa and DistilBERT. RoBERTA is a reimplementation of the classical BERT model with several minor tweaks to hyperparameters and embeddings. RoBERTa is a suitable choice for this study because it has been pre-trained on a corpus of Tweets, granting the model a more advanced understanding of Tweets and their lexicological components (Liu et al., 2019). Moreover, RoBERTa has been shown, in previous studies that used unimodal DL models to classify Pro-ED Tweets, to be the model with the highest performance overall (Benítez Andrades et al., 2022). DistilBERT is a distilled version of BERT and is approximately 40% smaller than the standard BERT model, allowing it to consistently attain faster training and inference speeds. DistilBERT's smaller size and condensed nature come with a trade-off: DistilBERT is regularly less accurate than RoBERTa or other vast BERT models (Sanh et al., 2020). DistilBERT is used in this study because it has been used in previous studies as a benchmark for NLP performance and compares the RoBERTa model (Benítez-Andrades et al., 2022).

Four DL models were chosen to constitute the multimodal model's CV segment: ViTB/16, ResNet-50, BEiT-L, and MaxViT. The ViT-B/16 is the base vision transformer model described earlier. It is included in the study because it has been employed in previous research examining CV models' capacity to identify Pro-ED posts (Feldman, 2023). The ResNet-50 model is the traditional CNN that was widely used and studied before the advent of vision transformers. Because of ResNet's past ubiquity, it is often used as a benchmark when investigating the performance of transformer-based models (Chen et al., 2022; Li et al., 2017). The BEiT model, which is an acronym for Bidirectional Encoder representation from Image Transformer, is a transformer-based model that differs from the standard vision transformer because it employs masking—the concealment of portions of the input image from the model—to improve the model's ability to generalize to new data (Bao et al., 2022). The MaxViT model is an updated version of the standard ViT model that combines both transformers and convolutions, thereby combining the superior qualities of CNNs and ViTs (Tu et al., 2022).

Each of the four CV models examined in this study was pre-trained on the ImageNet-1k dataset, which contains over one million data points and one thousand different data labels, also called classes (You et al., 2018). Among the four models, MaxViT is the newest and most accurate on the ImageNet-1k dataset. The second most accurate model is the BEiT-L model, the third is the ViT-B/16 model, and the fourth is the ResNet-50 model. Based on the results of previous studies that employed fine-tuned CV models, it is not guaranteed that models that have a higher accuracy on the ImageNet-1k dataset will be more adept at identifying Pro-ED images. Thus, it is pertinent to train and test multiple CV models (Fang et al., 2022).

## Procedure

Previous research has shown that DL models can generalize better when the dataset they are trained and evaluated on has an equal number of instances of all classes (Cao et al., 2022). Because of this, an under-sampling technique was employed to ensure that the minority class—the class with the fewest data points—was accurately represented. The minority class in the dataset was the Pro-ED class, with only 2,330 Tweets. Accordingly, the Pro-Recovery and the Neutral classes were randomly downsampled until they, too, had only 2,330 Tweets each, removing class imbalance from the dataset.

After the dataset was balanced, it contained 6,990 Tweets and was split into training, validation, and testing datasets. The training set comprised 60% of the dataset, while the validation and testing dataset comprised 20% each (Xu & Goodacre, 2018).

All image content within each dataset was fed through a series of transformations unique to each of the four CV models. These transformations are identical to those applied to the data fed into each model when originally trained (Hira & Gillies, 2015). Regardless of model type, all images were standardized through resized cropping and pixel normalization. Similarly, the text content from each Tweet was processed by a BERT tokenizer unique to each BERT model, which standardized the text and transformed it into a form processable by the models (Devlin et al., 2019). The labels, i.e., Pro-ED, Neutral, and Pro-Recovery, were given a unique number encoding—0, 1, and 2, respectively—so the models could process them.

The preprocessing measures were undertaken via PyTorch's *Torchvision* and *Dataset* packages, along with HuggingFace's *Transformer* package. Pre-trained RoBERTa, DistilBERT, MaxViT, ViT-B/16, and BEiT-L models were all acquired from HuggingFace's model repository, while the ResNet-50 model was instantiated through PyTorch's model library (Jiang et al., 2023; Paszke et al., 2019).

The four CV models were fine-tuned in conjunction with the DistilBERT and RoBERTa text processing models, generating eight trained models. Additionally, unimodal DistilBERT, RoBERTa, ViT, and MaxViT models were fine-tuned to be used as a benchmark in subsequent analysis. In total, twelve models were developed, four unimodal and eight multimodal.

# Results

| | MaxViT+ RoBERTa | MaxViT+ DistilBERT | ResNet-50+ RoBERTa | ResNet-50+ DistilBERT | BEiT-L+ RoBERTA | BEiT-L+ DistilBERT | ViT + RoBERTA | ViT + DistilBERT | RoBERTa | DistilBERT | MaxViT | ViT |
|---|---|---|---|---|---|---|---|---|---|---|---|---|
| Accuracy (%) | 95.9 | 89.6 | 92.5 | 87.8 | 93.1 | 89.2 | 92.8 | 88.5 | 88.3 | 85.5 | 79.3 | 77.4 |
| F1-Score | 0.959 | 0.896 | 0.925 | 0.878 | 0.931 | 0.891 | 0.928 | 0.885 | 0.883 | 0.854 | 0.792 | 0.774 |
| Precision | 0.959 | 0.897 | 0.925 | 0.878 | 0.931 | 0.891 | 0.927 | 0.888 | 0.884 | 0.854 | 0.793 | 0.776 |
| Recall | 0.959 | 0.895 | 0.925 | 0.878 | 0.931 | 0.891 | 0.928 | 0.885 | 0.884 | 0.855 | 0.793 | 0.774 |

**Table 1**: Accuracies, F1-Scores, Precisions, and Recalls for each model.

## Model Performance

Among the twelve models that were fine-tuned on the Tweet dataset, the model fusion of MaxViT and RoBERTa (MaxViT + RoBERTa), which is a novel fusion model first developed in this study was the best performing across all metrics. The MaxViT + RoBERTa model achieved an accuracy of 95.9% and an F1-score, precision, and recall of 0.959 when evaluated on the hold-out test dataset. These metrics, along with the metrics of the eleven other models, can be seen in **Table 1**.  Additional visual metrics for the MaxViT and RoBERTa fusion model, such as a confusion matrix, can be viewed in **Table 2**. The results of training and testing all twelve models align with the findings of previous studies in several crucial ways. First, multimodal models that employ transformer-based-image classifiers attain better results than multimodal models that employ CNNs (Cuenat & Couturier, 2022). Second, the multimodal fusion models that employ the RoBERTa model as the NLP section of their architecture, without fail, outperform in every metric the models that use DistilBERT. This is expected since RoBERTa was trained on a corpus of Tweets (Benítez-Andrades et al., 2022). Third, the multimodal models repeatedly achieve superior results than their unimodal models (Amal et al., 2022; Joze et al., 2020). Lastly, the unimodal NLP models—RoBERTa and DistilBERT—achieved comparable, albeit higher, accuracies and F1 scores to identical models developed in a previous study (Benítez-Andrades et al., 2022).

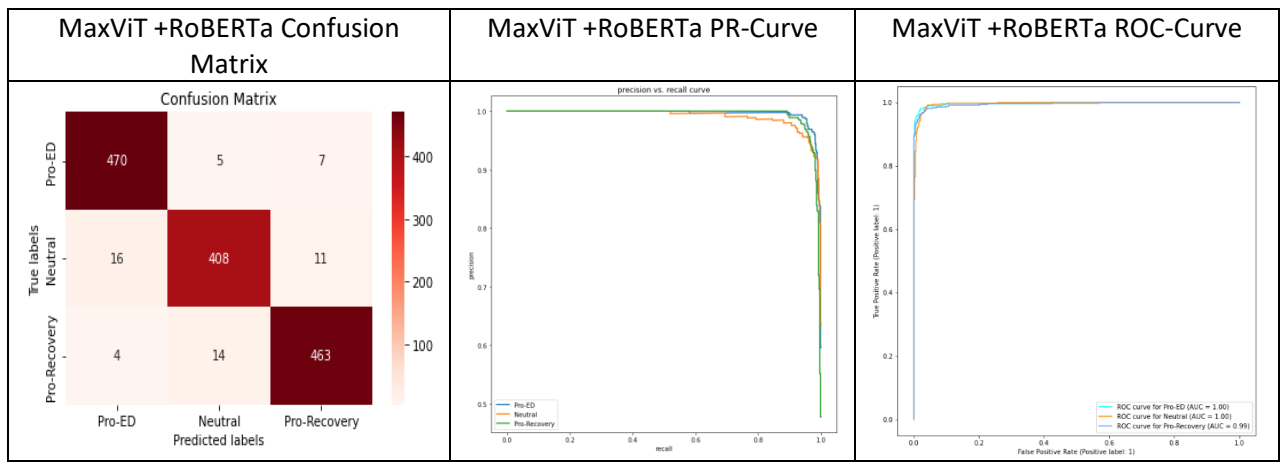

| MaxViT +RoBERTa Confusion Matrix | MaxViT +RoBERTa PR-Curve | MaxViT +RoBERTa ROC-Curve |
|---|---|---|

**Table 2**: Confusion Matrix, Precision vs. Recall Curve, and Receiver Operating Characteristic Curve for the

MaxViT and RoBERTa fusion model generated from the test dataset.

## Model Examination

After concluding that the MaxViT + RoBERTa model outperforms all other multimodal and unimodal models, a method to test the efficacy of the multimodal model on large social media datasets was devised. Tumblr hashtags that previous studies have shown to proliferate Pro-ED content were randomly sampled, starting from 2018 to 2023, using the Tumblr API (Branley & Covey, 2017; Chancellor et al., 2017). In total, 13,287 unique Pro-ED posts with both textual and visual data were scraped from the following hashtags: *#proana*, *#thinspiration*, *#thinspo*, *#bonespo*, *#collarbone*, and *#thighgap*—the same hashtags used to generate the Pro-ED Tweet dataset, though from a different social media platform.

The posts were preprocessed like the original Tweet dataset and fed into the MaxViT + RoBERTa generating a classification. According to the model, within the dataset of 13,287 Tumblr posts,

approximately 10,895 were of the Pro-ED class (~82.0%), 266 were of the Neutral class (~2.00%), and

2,126 were of the Pro-Recovery class (~16.0%). These findings are congruent with previous studies that used statistical modeling rather than DL to analyze the composition of Pro-ED hashtags on Tumblr (Branley & Covey, 2017).

In addition to Tumblr, several Pro-Recovery and ED-related Reddit subreddits—Reddit subcommunities created to discuss specific topics—were scraped using the Reddit API. The Pro-Recovery subreddits scraped were *r/fuckeatingdisorders*, *r/AnorexiaRecovery*, and *r/EatingDisorders*. Previous research has shown that these subreddits promote recovery from eating disorders (Fettach & Benhiba, 2020; Sowles et al., 2018). Indeed, the "About Community" biographies of each of the subreddits, which can be viewed in **Table 3**, professes that they are communities that promote eating disorder recovery. The ED-related subreddits were *r/EDanonymous* and *r/EDanonymemes*, which are sister subreddits that offer a judgment-free environment to share one's experiences with eating disorders. These two subreddits differ from the prior three because they do not actively promote recovery. Instead, they are a neutral forum for discussing ED-related topics (Feldhege et al., 2021; Kendal et al., 2017).

| r/AnorexiaRecovery | r/EatingDisorders | r/fuckeatingdisorders | r/EDanonymous | r/EDanonymemes |
|---|---|---|---|---|
| Sub for those trying to recover from Anorexia. No weights/numbers (calories) No personal information No before/after pics No specific behaviors No requests for "how to become anorexic" Message the mod with questions | r/EatingDisorders is a community dedicated to providing support, resources, and encouragement for individuals dealing with eating disorders. Whether you're in recovery, supporting a loved one, or seeking information, this subreddit is a supportive space with the aim to provide you with the support you need. | Eating disorders have many misconceptions, in part due to sufferers hiding their illness from loved ones who don't understand, perpetuating the cycle of silence. FED is here to confront eating disorders and provide a place for anyone to ask questions. | A public subreddit for discussing the struggles of having an eating disorder. Much like an Alcoholics Anonymous or Narcotics Anonymous group, we offer emotional support and harm reduction but no encouragement of furthering ED behaviors. This subreddit is not officially associated with the support group Eating Disorders Anonymous. We are not exclusive to or trying to "force" recovery on anyone. | Welcome to r/EDanonymemes - the chaotic sister sub of r/EDanonymous! Much like other depression meme subreddits, this is a supportive space for people with eating disorders to share relatable memes and cope with dark humored shitposting. We do not encourage self-harm or tolerate any pro-ana content. We are not exclusive to or trying to "force" recovery on anyone. Meme without judging each other! |

**Table 3:** "About Community" biographies for five Reddit subreddits, each provided by the subreddit moderators.

The purpose of analyzing the MaxViT + RoBERTa model's performance on Reddit posts is to measure its capacity to examine content from a social media site significantly different from the one it was trained on: Twitter (Sowles et al., 2018). On Reddit, Pro-ED communities are moderated far more strictly than on Twitter or Tumblr, and the foremost Pro-ED subreddit on the platform, *r/proED*, was removed from the site several years ago, shattering the Pro-ED community on Reddit (Chancellor et al., 2018; Feldhege et al., 2021). Thus, unlike Twitter or Tumblr, Reddit no longer has recognizable or researched Pro-ED communities.

Overall, 11,918 Reddit posts were randomly sampled from the Pro-Recovery subreddits, 447 of which contained both visual and textual content, and 6,071 Reddit posts were randomly sampled for the ED-related subreddits, 3,728 of which contained visual and textual content. Only the posts containing visual and textual inputs were fed into the MaxViT + RoBERTa model for examination. Of the 447 posts sampled from Pro-Recovery subreddits, the model identified 128 (~28.6%) as Pro-ED, 18 as Neutral (~4.02%), and 301 (~67.3%) as Pro-Recovery. Among the 3,728 ED-related posts sampled, the model identified 2,916 (~78.2%) as Pro-ED, 111 as Neutral (~2.98%), and 701 (~18.8%) as Pro-Recovery. The results of the analysis of the Pro-Recovery subreddits align quantitatively with previous studies (McCaig et al., 2018, 2020) and are supported by previous research that found Pro-Recovery communities on social media sites often contain a host of Pro-ED content (Chancellor, Mitra, et al., 2016; Fettach & Benhiba, 2020; Jones et al., 2022). Similarly, the results of the analysis of the ED-related subreddits are akin to previous studies that found anonymous ED discussion forums to be rife with Pro-ED content (Chung et al., 2021; Feldhege et al., 2021; Sidani et al., 2016).

The similarity between the findings of previous studies (Fettach & Benhiba, 2020; McCaig et al., 2018; Sidani et al., 2016) and the results of the MaxViT + RoBERTa model's classifications, when fed a dataset of randomly sampled Tumblr and Reddit posts, indicates that the model can generate classifications that are representative of previous research and that the model, even though it was trained solely on Twitter data, can generalize to other social media sites (López-Vizcaíno et al., 2023).

## Social Media Analysis

After verifying the MaxViT + RoBERTa model on Pro-ED Tumblr posts and ED-related Reddit Posts, a comprehensive social media content analysis was undertaken to examine the fluctuations in the relative abundance of Pro-ED content on Twitter over time. Overall, eight popular Twitter hashtags were chosen. For this study, Twitter hashtags were considered popular if they have, on average, over five thousand Tweets posted each week in the year 2023. These hashtags fell into two groups: Twitter hashtags previous studies revealed to contain posts like those found in Pro-ED communities, and hashtags that contain keywords that previous studies have shown to be highly prevalent in the Pro-ED community within their names. The hashtags that contained posts akin to Pro-ED hashtags, as described in previous research (Chancellor et al., 2017; S. N. Counts et al., 2018; Feldman, 2023), were *#selfie, #ootd*, *#model*, and *#fashion*.  The hashtags that contained keywords associated with Pro-ED content in their names were: *#weightloss*, *#fitnessmotivation*, *#healthyliving*, and *#healthylifestyle*. This analysis aimed to track the fluctuations of Pro-ED content on the hashtags over many years (Nguyen et al., 2012).

To this end, the previously mentioned hashtags were scraped via stratified random sampling. The stratified random sampling method used in this study adhered to the following procedure. Three random non-consecutive days were selected for any given month within the sampling period, and all Tweets from that day were scraped. The posts with both visual and textual content were extracted from these Tweets to be loaded into the MaxViT + RoBERTa model.

The above sampling process was repeated for each of the eight hashtags from January 2014 to April 2023—equating to one hundred and twelve months of data. The data for each hashtag was aggregated by month and subsequently fed into the model, generating classifications. Based on the percent of Tweets classified as Pro-ED by the model for any given month when compared to the total number of examined Tweets for said month, a relative abundance value was generated. These relative abundance values were then arranged by hashtag and graphed. Additionally, third-degree polynomial regressions were fit to each graph of Pro-ED content relative abundance, all attaining a P-value of less than 0.001 (P < 0.001). The graphs of the percent composition of Pro-ED content and polynomial regressions for the four hashtags from previous studies can be viewed in **Table 4**. The same data for the hashtags derived through keyword similarity can be viewed in **Table 5**.

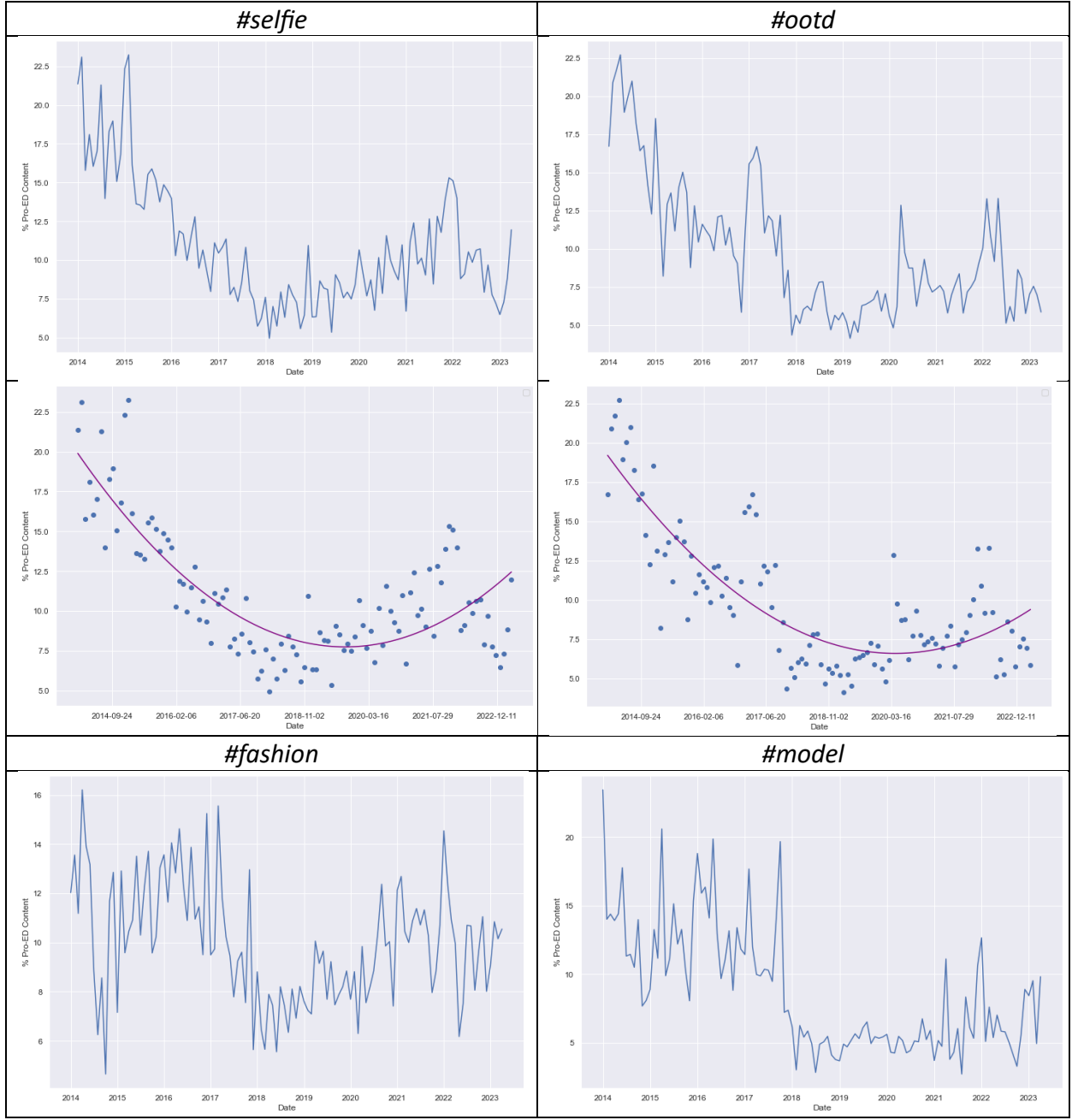

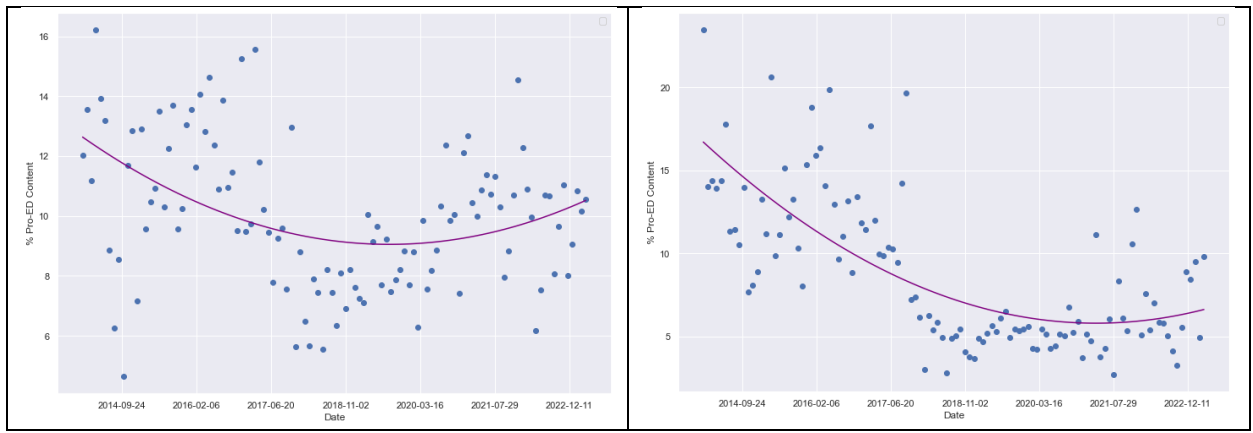

**Table 4**: Graphs illustrating the relative abundance, in percent, of Pro-ED content on the four hashtags mentioned in (Chancellor et al., 2017; S. N. Counts et al., 2018; Feldman, 2023), over the period from January 2014 to April 2023, along with graphs displaying a third-degree polynomial regression fit to that data.

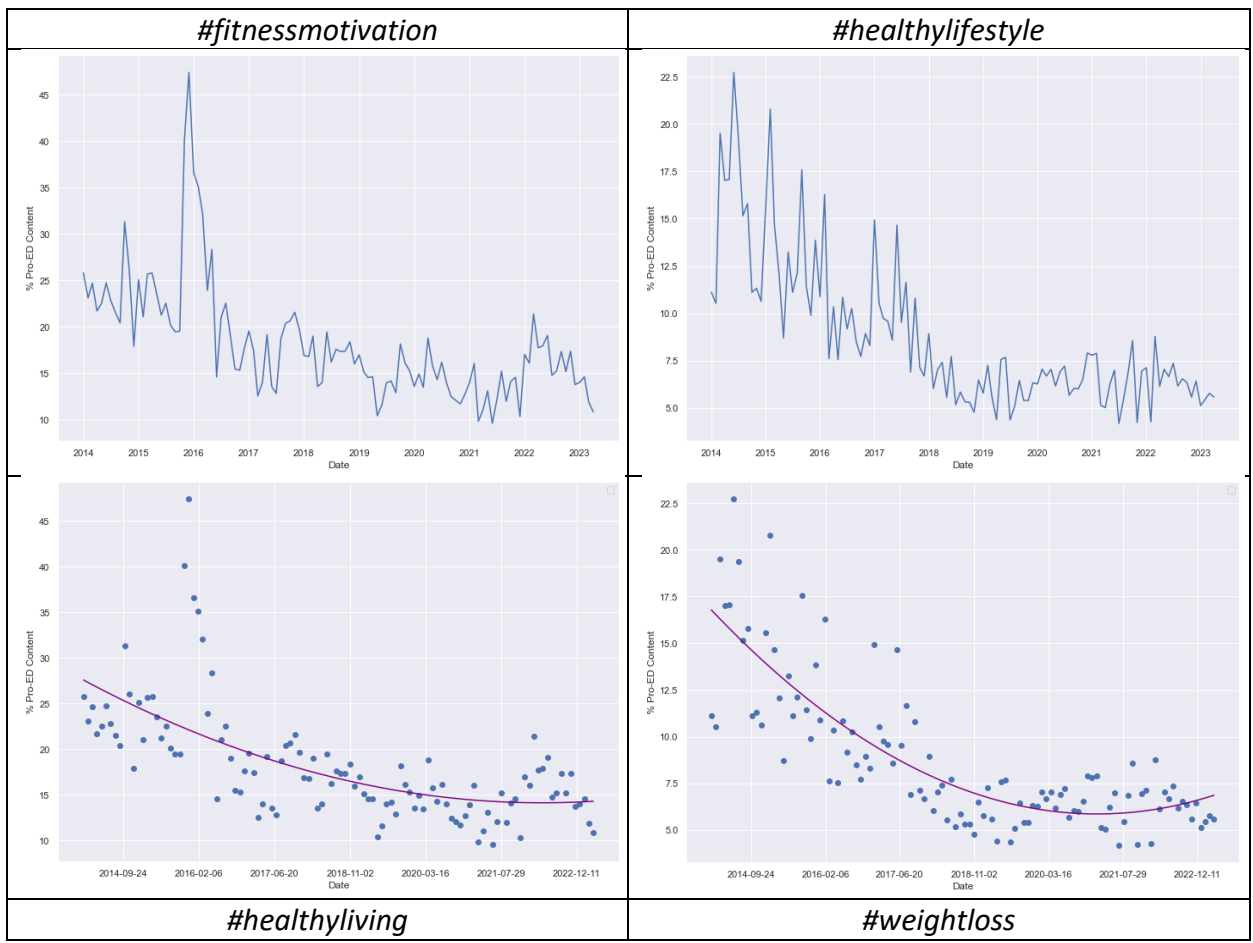

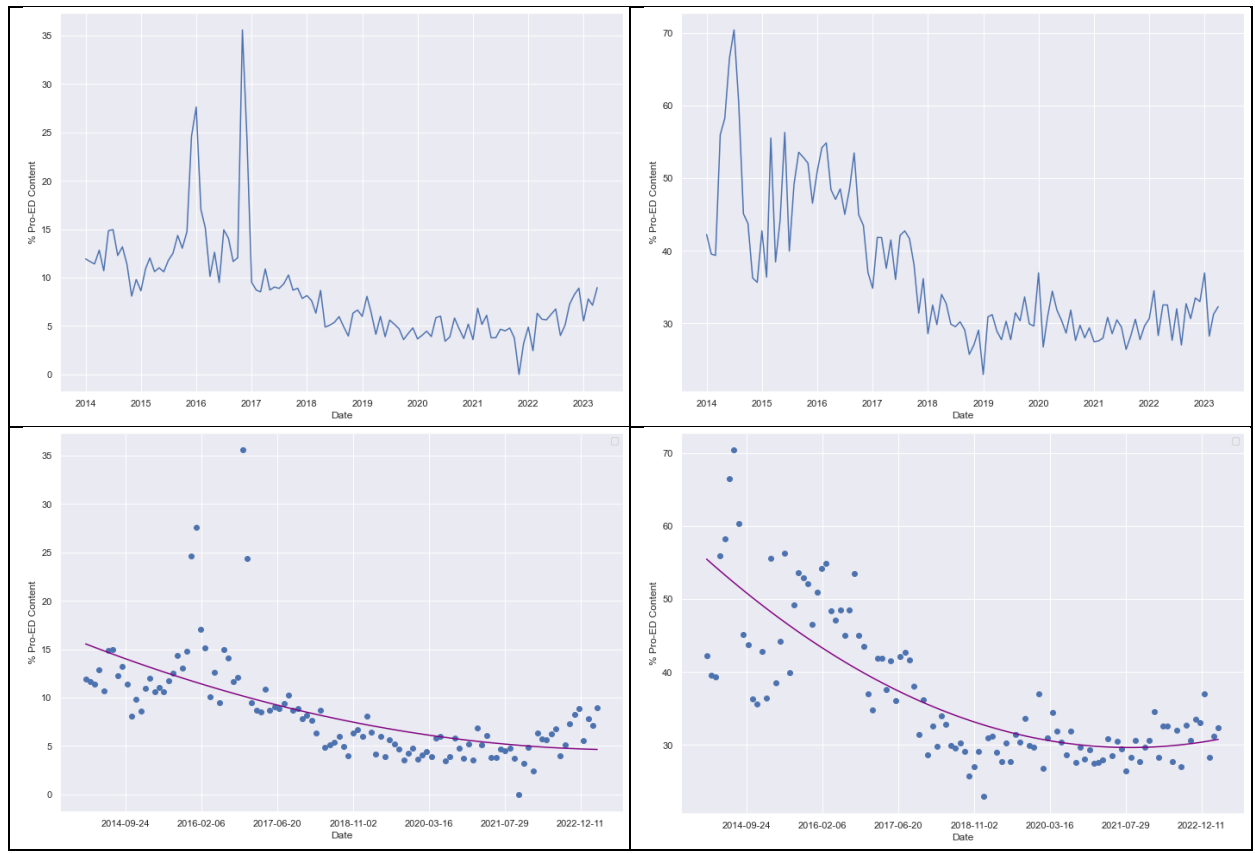

**Table 5**: Graphs illustrating the relative abundance of Pro-ED content, in percent, on the four hashtags that share keyword similarities with Pro-ED communities on Twitter, over the period from January 2014 to April 2023 along with graphs displaying a third-degree polynomial regression fit to that data.

The data extracted from the eight Twitter hashtags suggest that, over the last ten years, there has been a drastic decrease in the relative abundance of Pro-ED content on Twitter. For most of these hashtags, the relative abundance of Pro-ED content decreased consistently until late 2018 and early 2019. Afterward, the relative abundance of Pro-ED content either entered a state of equilibrium or began to rise again, as seen with the four hashtags noted by previous studies: *#selfie*, *#fashion*, *#ootd*, and *#model*. To visualize these trends, the dataset was adjusted to exclude any data points from before 2018, and then, a linear regression was fit to the modified dataset—the results of which can be seen in

**Table 6**. These findings align with a prior study (Feldman, 2023), which found that Pro-ED content on *#selfie* had increased since 2018, though that study used unimodal DL models.

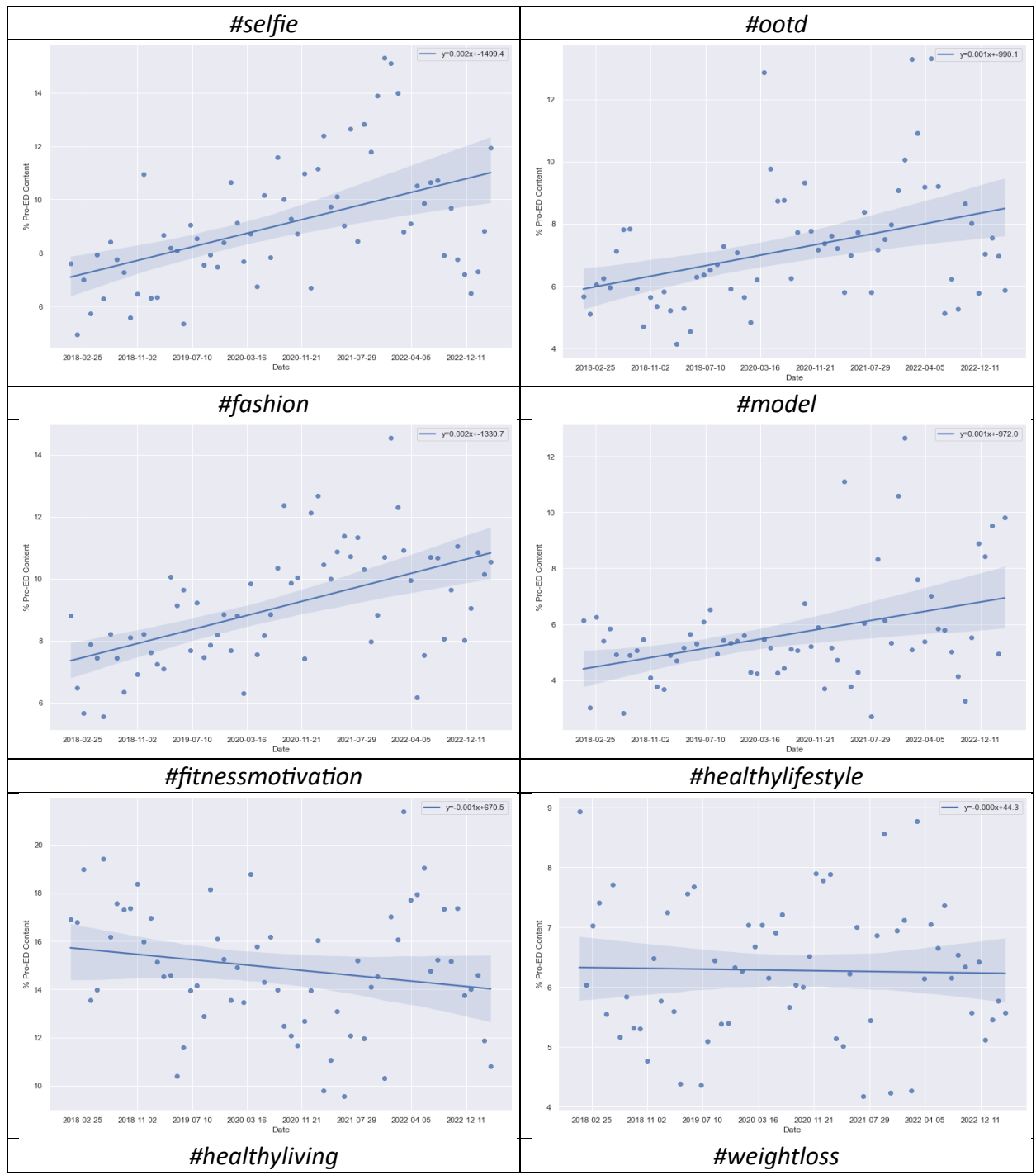

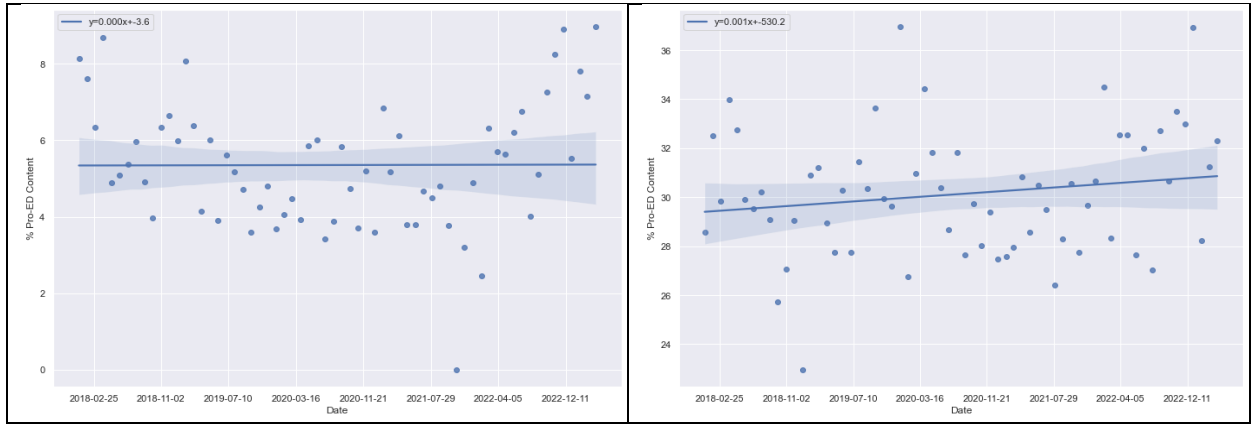

**Table 6**: Graphs of linear regressions for all datasets of relative abundance of Pro-ED content, in percent, sampled from January 2018 to April 2023.

The data extracted from the hashtags stipulates that, though it had become less copious on Twitter leading up to 2018, Pro-ED content has rebounded, no longer shrinking, or growing in abundance across hashtags. Moreover, according to the linear regressions, the growth in Pro-ED relative profusion will continue in the future, possibly to its previous ubiquity.

# Discussion

## Principal Findings

This is the first study to explore the efficacy of developing and deploying multimodal DL models to classify Pro-ED content on social media. Twelve multimodal NLP and CV fusion DL models were trained on a dataset of Tweets encompassing multiple modalities procured from Twitter. They were then assessed to arbitrate their accuracy on an unseen test dataset. The results of this assessment indicated that all twelve models were effective at recognizing Pro-ED, Neutral, and Pro-Recovery content—the lowest accuracy and F1 score of any model was 77.4% and 0.774, respectively, while the highest 95.9% and 0.959, respectively. Despite the efficacy of all the DL models developed, several were hindered either by their unimodality or the NLP unit they possessed. Models that employed only one modality for classification—image or text—performed significantly poorer than their multimodal counterparts.

Additionally, multimodal models that employed a RoBERTa NLP unit performed significantly better than models that utilized DistilBERT. These outcomes were expected as many previous studies have indicated that multimodal models outperform unimodal models consistently and that, when classifying social media posts, especially Tweets, RoBERTa outperforms DistilBERT (Benítez-Andrades et al., 2021; Boulahia et al., 2021). The accuracy and F1-score of all multimodal models developed in this study outperform any unimodal method for detecting Pro-ED content on Twitter. There have yet to be other studies to classify Pro-ED content on social media using multimodal DL, so a comparison between performance cannot be drawn.

After training and assessing the twelve DL models, the MaxViT + RoBERTa fusion model, a novel model first developed in this study, was determined to be the most accurate. The MaxViT + RoBERTa was deployed to classify a vast dataset of ED-related posts collected from Tumblr and Reddit, which no study has yet done. The examination revealed that Pro-ED communities on Tumblr contain sizeable portions of Pro-Recovery content. Additionally, by identifying very few posts as Neutral, the model could distinguish content that did not contain eating disorder valence, whether advocacy or derision. Of course, there will likely be Neutral posts on Pro-ED hashtags because social media hashtags have no content requirements (La Rocca & Boccia Artieri, 2022). The analysis also revealed that Pro-Recovery and eating disorder support communities on Reddit contained much Pro-ED content. These findings are congruent with the conclusions of previous studies about the composition of Pro-ED and Pro-Recovery hashtags on Tumblr and Reddit (Branley & Covey, 2017; Fettach & Benhiba, 2020; McCaig et al., 2018). Unlike this study, those studies employed statistical and qualitative approaches rather than AI. The similarity between the MaxViT + RoBERTa model's classifications and previous research evidences the efficacy of the model and the extent of its insight into Pro-ED content.

Next, the MaxViT + RoBERTa fusion model was deployed to analyze Tweets scraped through a stratified random sampling methodology. The collected Tweets were arranged by month and graphed. These graphs then had polynomial regressions, which can be viewed in **Table 4** and **Table 5**, fit to them. According to these graphs, the relative abundance of Pro-ED content across all eight hashtags has been decreasing rapidly since 2014, a trend yet undocumented by other studies. Furthermore, six of the eight hashtags experienced the lowest relative abundance of Pro-ED content between January 2018 and December 2019. The reason behind this sudden and unforeseen depletion of Pro-ED content on these Twitter hashtags, and, likely, across many Neutral Twitter hashtags, is yet to be studied. A possible explanation for this decrease is increased moderation and removal of content promoting self-harm or provocative practices on Twitter, which culminated in an overhaul of the Twitter content rules and guidelines, explicitly banning content promoting self-harm, like eating disorders (Shu, 2018; Twitter's Policy on Hateful Conduct | Twitter Help, n.d.). Another possible explanation is that the Pro-ED community on Twitter had dispersed, leaving the platform in search of other social media sites. The year 2014, when the stark drop in Pro-ED relative abundance began, was the same year that the social media application Musical.ly was released, which became an instant success, surpassing all other social media platforms in popularity on the IOS popularity list (Carson, 2016). In 2018, Musical.ly was bought by ByteDance Ltd and rebranded as TikTok (Kaye et al., 2022).

Similarly, first launched in 2010, Instagram had amassed over two hundred million users by 2014 and breached the one billion user threshold by 2018 (Chancellor, Pater, et al., 2016; Jensen, 2013). TikTok and Instagram are used daily by 58% and 50% of teenagers, aged 13 to 17, respectively (Vogels & Gelles-Watnick, n.d.), the age demographic when ED onset is highest and Pro-ED content attunement is most likely. It is also possible that the cause of the decline in Pro-ED content is a combination of the factors above.

What is most striking, however, is that Pro-ED content has either stopped its decline or rebounded in percent abundance in seven out of the eight sampled hashtags. Pro-ED content is again growing on these hashtags and, possibly, across Twitter. The likely cause of this sudden increase is the Covid-19 pandemic, which previous studies found to have enlarged the average time spent on social media and the prevalence of Pro-ED content (Agostino et al., 2021; Khosravi, 2020, 2020). Based on the linear regressions generated from classified Tweets, the relative abundance of Pro-ED content will likely keep growing, reinvigorated by the Covid-19 pandemic.

## Limitations

There are several limitations within the study conducted. First, the Twitter dataset upon which deep learning models were trained in this study contained only multimodal Tweets, thus not including posts that are solely textual or visual. Second, the dataset on which the DL models were trained consisted of data scraped from hashtags that previous studies have analyzed, therefore only encompassing a portion of the Pro-Eating Disorder and Pro-Recovery movements on social media. Third, the examinations of Tumblr, Reddit, and Twitter described in this paper only analyzed data that included both textual and visual modalities.

# Conclusion

Multimodal deep learning models were trained to classify whether a post, consisting of textual and visual elements, promoted eating disorders, encouraged informed recovery from eating disorders, or was unrelated to eating disorders and the previous two post types. Twelve deep learning models were trained, all achieving accuracies greater than 77%. The best performing model was the multimodal fusion of the RoBERTa NLP model and the MaxViT model, which attained an accuracy of 95.9%

The RoBERTa and MaxViT fusion model was then evaluated on a dataset of randomly sampled posts from Pro-Eating Disorder Tumblr communities and eating disorder-related Reddit subreddits. The results of this examination were consistent with those of previous such studies, showing that the fusion model can replicate the outcomes of previous research. Afterward, the RoBERTa and MaxViT fusion model was used to analyze eight Twitter hashtags, uncovering that, over the last decade, Pro-Eating Disorder content has decreased in relative abundance on Twitter, reaching its lowest threshold in 2018. The causes of this decrease are unknown, and more research is required to understand this phenomenon better. Additionally, the analysis revealed that after 2018, many of the hashtags had a resurgence in Pro-Eating Disorder content. This revival coincided with the start of the Covid-19 pandemic, divulging the profound effect of the Covid-19 pandemic and associated stressors on the prevalence of Pro-ED content on social media.

Future research will focus on increasing the modalities of analysis, creating deep learning models capable of analyzing video, and utilizing the trained fusion models on real-time social media data, potentially as an autonomous bot, to identify and report posts and platforms that promote eating disorders online.